\documentclass[runningheads]{llncs}

\usepackage{graphicx}
\usepackage{tikz}
\usepackage{comment}
\usepackage{amsmath}
\usepackage{amssymb}
\usepackage{pgfplots}
\usepackage{subfigure}
\usepackage{color}
\usepackage[accsupp]{axessibility}  
\usepackage{multirow}
\usepackage{booktabs}
\usepackage[pagebackref]{hyperref}
\newcommand{\dx}[1]{~\mathrm{d}#1}
\newcommand{\idest}{\textit{i.e.},~}

\newcommand{\etal}{\textit{et al}.~}
\providecommand{\viz}{\textit{viz.,}~}

\DeclareMathOperator*{\argmin}{arg\,min}

\usepackage[capitalize,noabbrev]{cleveref}
\usepackage{orcidlink}

\pgfplotsset{compat=1.18}
\title{
  LocaliseBot: Multi-view 3D object localisation with differentiable rendering
  for robot grasping
}

\begin{document}
  \mainmatter
  \def\ECCVSubNumber{030}

  \titlerunning{LocaliseBot}

  \author{
      Sujal Vijayaraghavan\inst{1,3}\orcidlink{0000-0003-2716-8199} \and
      Redwan Alqasemi\inst{2,3} \and
      Rajiv Dubey\inst{2,3} \and
      Sudeep Sarkar\inst{1,3}\orcidlink{0000-0001-7332-4207}
  }

  \authorrunning{S. Vijayaraghavan \etal}

  \institute{Department of Computer Science \and
    Department of Mechanical Engineering \and
    University of South Florida, Tampa\\
    \email{\{sujal,alqasemi,dubey,sarkar\}@usf.edu}\\
  }

  \maketitle

  \begin{abstract}
    Robot grasp typically follows five stages: object detection, object
    localisation, object pose estimation, grasp pose estimation, and grasp
    planning. We focus on object pose estimation. Our approach relies on three
    pieces of information: multiple views of the object, the camera's extrinsic
    parameters at those viewpoints, and 3D CAD models of objects. The first step
    involves a standard deep learning backbone (FCN ResNet) to estimate the
    object label, semantic segmentation, and a coarse estimate of the object
    pose with respect to the camera. Our novelty is using a refinement module
    that starts from the coarse pose estimate and refines it by optimisation
    through differentiable rendering. This is a purely vision-based approach
    that avoids the need for other information such as point cloud or depth
    images. We evaluate our object pose estimation approach on the ShapeNet
    dataset and show improvements over the state of the art. We also show that
    the estimated object pose results in 99.65\% grasp accuracy with the ground
    truth grasp candidates on the Object Clutter Indoor Dataset (OCID) Grasp
    dataset, as computed using standard practice.
\end{abstract}

  \section{Introduction}
\label{sec:introduction}

\begin{figure}[htb!]
    \centering
    \subfigure[]{
        \includegraphics[scale=0.14]{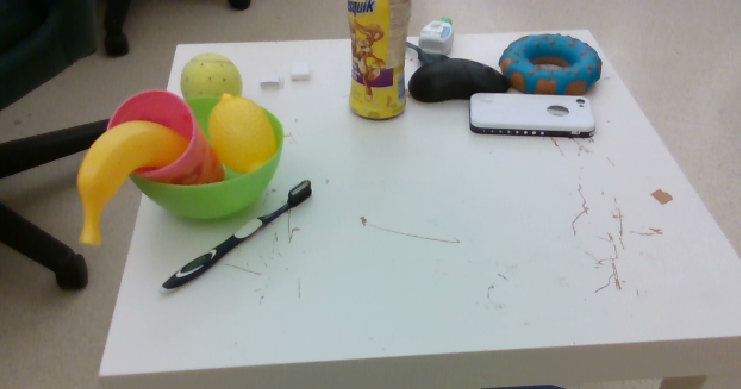}
        \label{subfig:depth1}
    }
    \subfigure[]{\includegraphics[scale=0.12]{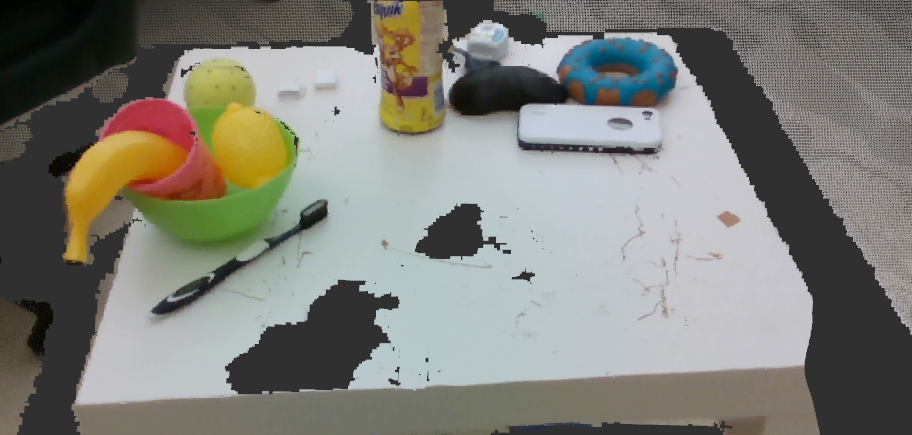}}
    \subfigure[]{
        \includegraphics[scale=0.1]{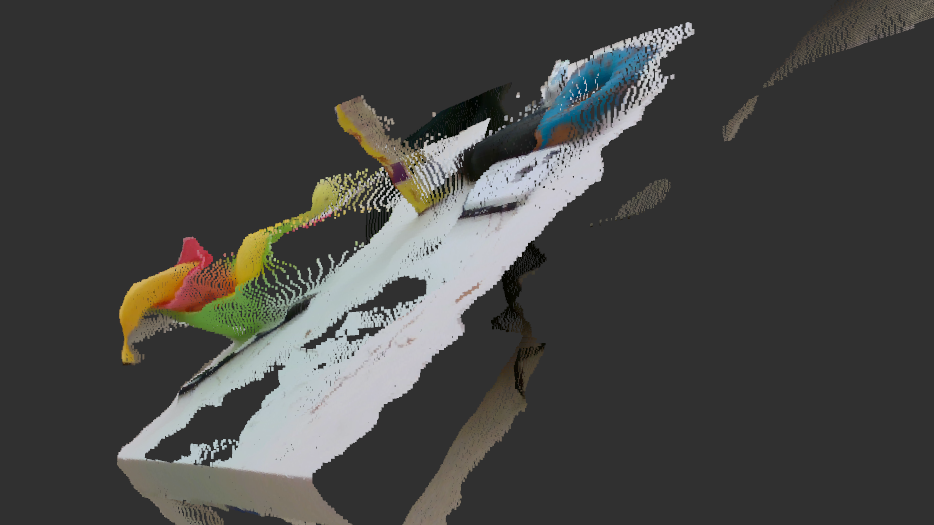}
        \label{subfig:depth3}
    }
    \caption{
        RGB image (a) of a scene, (b) its corresponding point cloud, and (c) and
        the point cloud from a different viewpoint. These images illustrate how
        point cloud/depth sensors can lose significant amounts of data from an
        original scene resulting in unstable performance of models relying on
        them.
    }
    \label{fig:depth-drawbacks}
\end{figure}

The problem of grasp pose estimation is more challenging for gripper-based end
effectors compared to suction-based ones. The former requires a good
understanding of the object shape before grasping. Point cloud can provide
valuable depth information, but methods that rely on depth sensor data such as
point cloud information often suffer from missing or noisy data (see
\cref{fig:depth-drawbacks}).

Most grasp detection techniques operate in the pixel space. Typically, they rely
on object regression followed by grasp pose regression. The estimations include
detection of the object category, semantic segmentation, and for pose estimation
for robot grasping, an algorithm that proposes grasp candidates. The regression
lacks the crucial depth information. This results in a vast grasp candidate
space to optimise. Other robot grasping techniques
\cite{chen2021fs,le20216d,huang2021comprehensive} utilise other available
sensory information such as point cloud, for example.

A 3D awareness of the objects can significantly reduce the search space for
grasp pose candidates. This paper proposes utilising available knowledge of
geometric shapes and properties of objects detected in the form of CAD models.

\subsection{3D model fitting}

\begin{figure}[htb!]
    \centering
    \begin{tabular}{@{}c@{}}
        \includegraphics[scale=0.4]{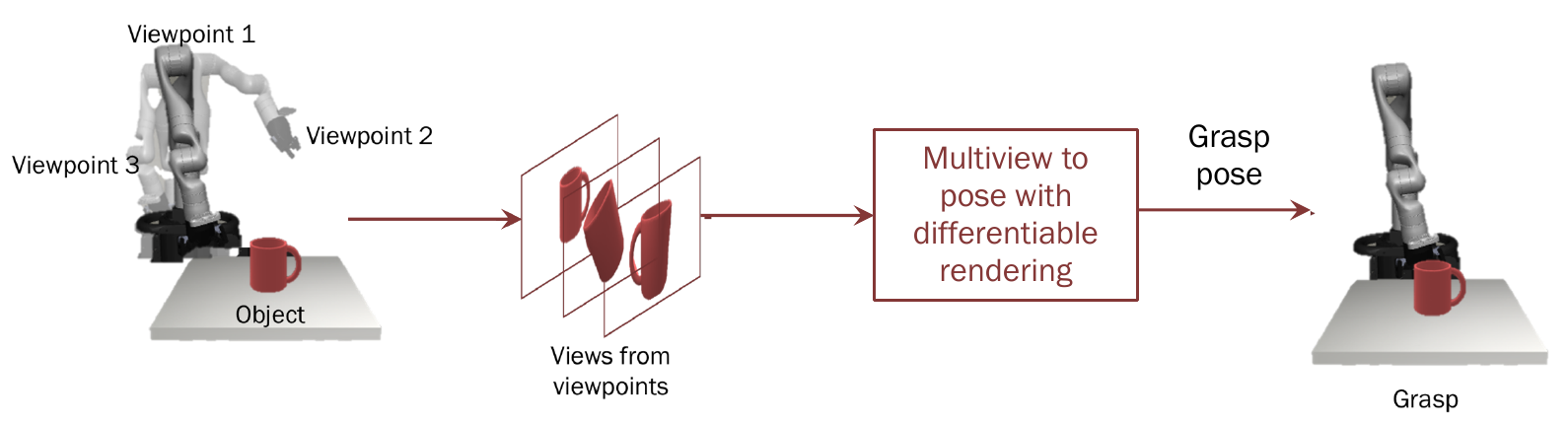}
        \\[\abovecaptionskip]
        \small (a) Arm manipulation to capture the scene from multiple
        viewpoints
    \end{tabular}

    \begin{tabular}{@{}c@{}}
        \includegraphics[scale=0.55]{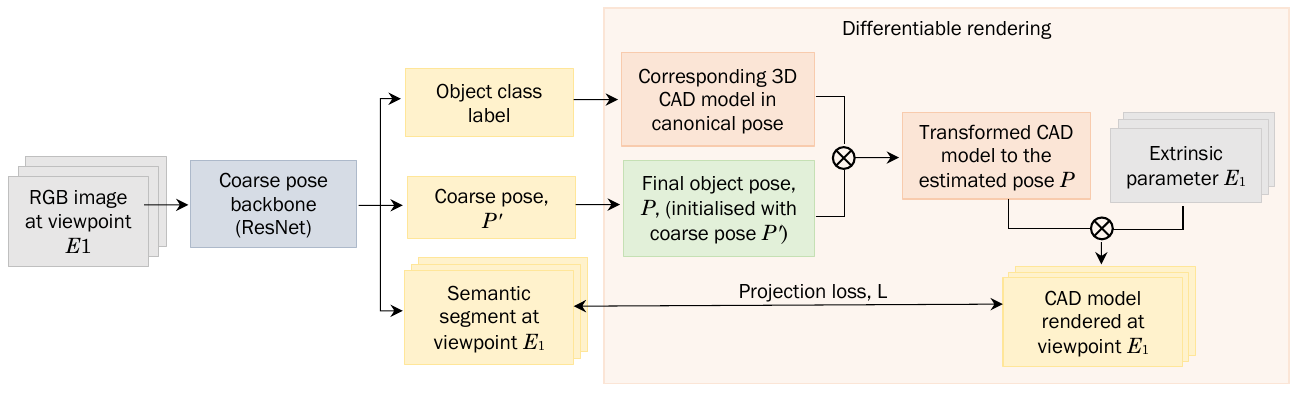}
        \\[\abovecaptionskip]
        \small (b) Pipeline of the method
    \end{tabular}

    \caption{
        An overview of the proposed approach which includes (a) capturing the
        scene from multiple viewpoints and (b) estimating the object pose from
        the multiple views. Multi-view pose refinement is achieved through
        differentiable rendering by accepting the multiple camera parameters, a
        3D CAD model of the detected object, and projects them with a shared set
        of object pose parameters. The rendering at each viewpoint is compared
        with the observed image and the error is backpropagated to optimise the
        shared object pose parameters.
    }
    \label{fig:overview}
\end{figure}

One of the primary challenges facing the task of 3D model alignment is the
estimation of depth information of objects from 2D images. This is because the
depth estimation problem lacks sufficient constraints from a single 2D image and
consequently suffers from the \textit{depth-scale ambiguity}, a long-time
challenge tackled to date since the classical computer vision era.

In classical computer vision, camera calibration techniques
\cite{song2013survey} estimate structure from motion. Such methods generally
rely on three pieces of information: images of a scene from multiple viewpoints,
camera parameters, and mapping of keypoints of an object between the multiple
viewpoints. While the former two artefacts are easy to obtain, the latter is
not, especially for real-time applications.

More recently, deep learning models are trained to directly estimate the 3D pose
of objects in images.\cite{kuo2020mask2cad,gkioxari2020mesh} Such models are
trained on large datasets with the depth information annotated. They do not
require any annotation at inference time. However, estimation of depth from a
single viewpoint---even with a fully-trained network---is often inaccurate and
unstable. Pose estiation from multiple views are now being applied to refine the
initial estimates.\cite{maninis2022vid2cad}

In this work, we use the modified FCN ResNet \cite{long2015fully} initialised
with pre-trained weights and fine-tune it for coarse pose estimation. This,
along with extrinsic camera parameters, are used for multiview pose refinement.
Extrinsic camera parameters are often available in robot grasping problem
scenarios. In order to reduce the high search space for the pose, we break down
the search space into bins and, from them, select the ones with the best
matching (\cref{sec:search-space-reduction}). For run-time inference, multiple
views in an explorative fashion are obtained (\cref{fig:overview}a).

This method relies on RGB images, extrinsic camera parameters, and a knowledge
base (of CAD models) and does not use point cloud or other depth sensor
information, making it relevant to applications with simple RGB sensors. The
overall pipeline is illustrated in \cref{fig:overview}b.

  \section{Related work}

Object pose estimation for robot grasping typically goes through three stages:
object localisation, object pose estimation, and grasp pose
estimation.\cite{du2021vision} Our focus in this work is on object pose
estimation and refinement. Existing methods
\cite{kumra2020antipodal,cheron2015p,li2019cdpn,li2020pose} combine object
localisation and object pose estimation into a single system (of one or more
modules) and tackle the whole problem. For grasp pose estimation, some methods
\cite{vohra2019real,liu2020grasp,wei2022dvgg} rely on point clouds in addition
to image features, whereas others use depth
images.\cite{supancic2015depth,bai2021active,litvak2019learning,buchholz2013efficient,wang2019densefusion}
In recent computer vision techniques, 3D object pose estimation is done through
deep learning. Some techniques estimate object pose based on point cloud
\cite{li2020category} or depth images.\cite{wang2019normalized}

Existing methods rely on a variety of available modes of data and effectively
estimate grasp pose estimation. Deep learning models
\cite{xiang2017posecnn,tekin2018real,ainetter2021end} are trained to implicitly
estimate the object pose. An iterative inference step as a downstream task over
learnt deep models is also applied to further improve estimation
accuracy.\cite{wu2020deep} A recent review \cite{du2021vision} recounts the
literature based on object localisation, pose estimation, and grasp estimation.

\subsection{3D pose estimation}

Numerous techniques exist for various degrees of pose
estimation.\cite{li2018deepim,pitteri20203dobject} Different techniques have
been applied depending on available data and modes.

More recent works rely on established detection models and built on them. For
example, Mesh-RCNN \cite{gkioxari2020mesh} uses Mask-RCNN \cite{he2017mask} as a
backbone to predict a coarse voxel representation of the detected object.
Further refinement on it is achieved by learning a graph neural network to infer
the shape.

ShapeMask,\cite{ku2019shapemask} augmenting on ResNET,\cite{he2016deep} learns
to output a feature vector for the object detected. This end-to-end architecture
learns to detect objects and output a vector summarising geometric features of
the object.

Mask2CAD \cite{kuo2020mask2cad} learns to map the latent vector of object
features generated by ShapeMask to a latent representation of 3D CAD models. In
effect, the model can detect objects and retrieve an appropriate 3D CAD model. A
third head is trained also to predict the object pose.

Depth estimation from a single image, however, is often unstable and less
reliable. Video2CAD \cite{maninis2022vid2cad} builds on Mask2CAD to obtain a
coarse pose of the object, extending the idea to multiple views in a video.
Explicit constraints for localisation and scaling of the object are placed
across image frames and optimised for the overall videoframes. Multiple views
with sufficient distinction provide the necessary information to deduce scale
and depth.

\subsection{Differentiable rendering}
\label{sec:differentiable-rendering}

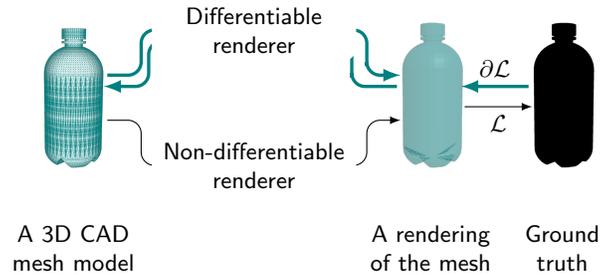
\begin{figure}[tb!]
    \centering
    \begin{tikzpicture}[scale=0.6]
        \node (mesh) at (0, 0) {
            \includegraphics[scale=0.05]{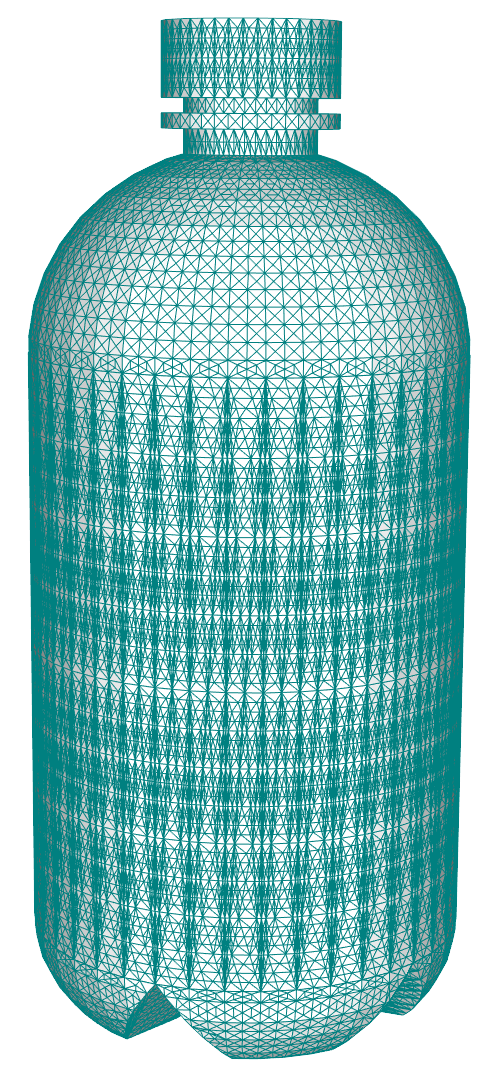}
        };

        \node[below of=mesh, node distance=2cm,text width=1.75cm,
            text centered] {
            \sffamily{A 3D CAD mesh model}
        };

        \node[right of=mesh, node distance=4.75cm] (rendering){
            \includegraphics[scale=0.1]{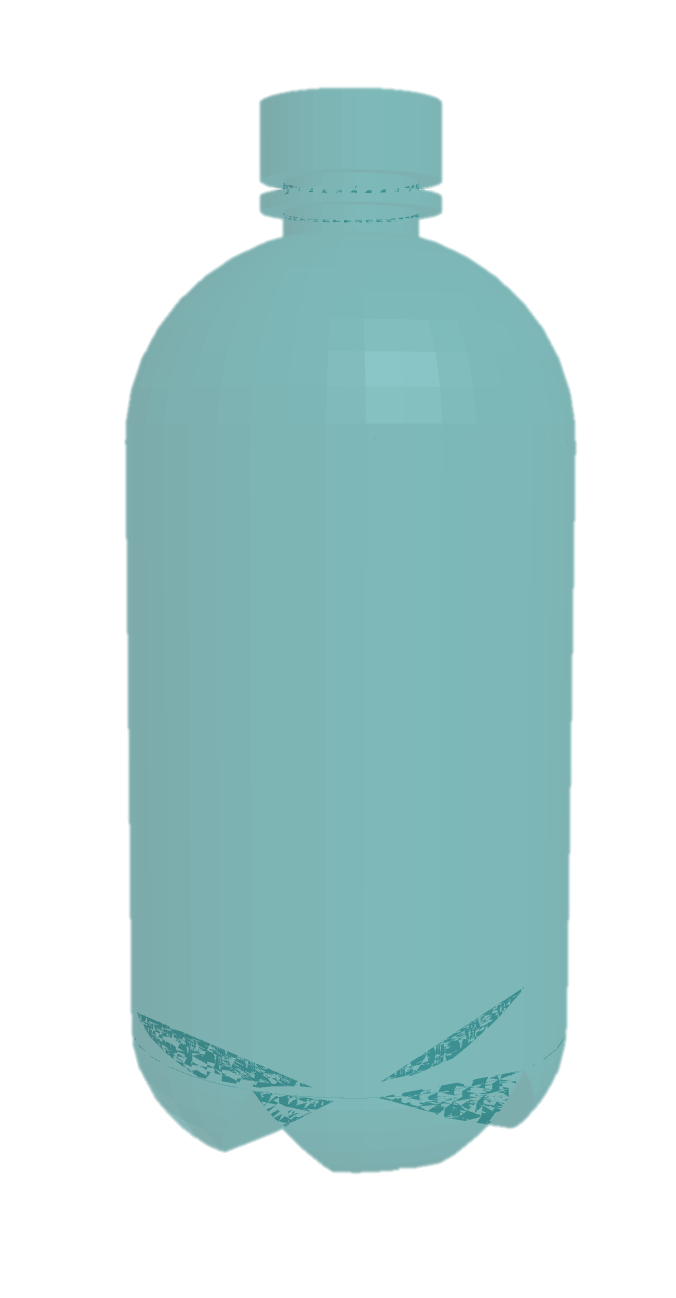}
        };

        \node[below of=rendering, node distance=2cm, text width=1.75cm,
            text centered] {
            \sffamily{A rendering of the mesh}
        };

        \draw[-latex, line width=1.25pt, rounded corners=5pt, draw=teal]
            (rendering) ++(-0.75, 0.25) -- ++(-1, 0) -| ++(0, 1) --
            ++(-4.5, 0) -- ++(0, -1) -- ++(-1,0) (mesh);

        \draw[-latex, line width=1.25pt, rounded corners=5pt, draw=teal]
            (mesh) (0.75, 0.5) -- ++(0.75,0) -- ++(0, 1) -- ++(5, 0)

            node[midway, rectangle, fill=white, text width=2.5cm,
                text centered]{
                \sffamily{Differentiable renderer}
            } -- ++(0, -1) |- ++(0.75, 0) (rendering);
        
        \draw[-latex, rounded corners=5pt] (mesh) (0.75, -0.5) --
            ++(0.75,0) -- ++(0, -1) -- ++(5, 0) node[midway, rectangle,
            fill=white, text width=2.5cm, text centered]{
                \sffamily{Non-differentiable renderer}
            } -- ++(0, 1) -- ++(0.75, 0) (rendering);

        \node[right of=rendering, node distance=1.75cm] (gt) {
            \includegraphics[scale=0.1]{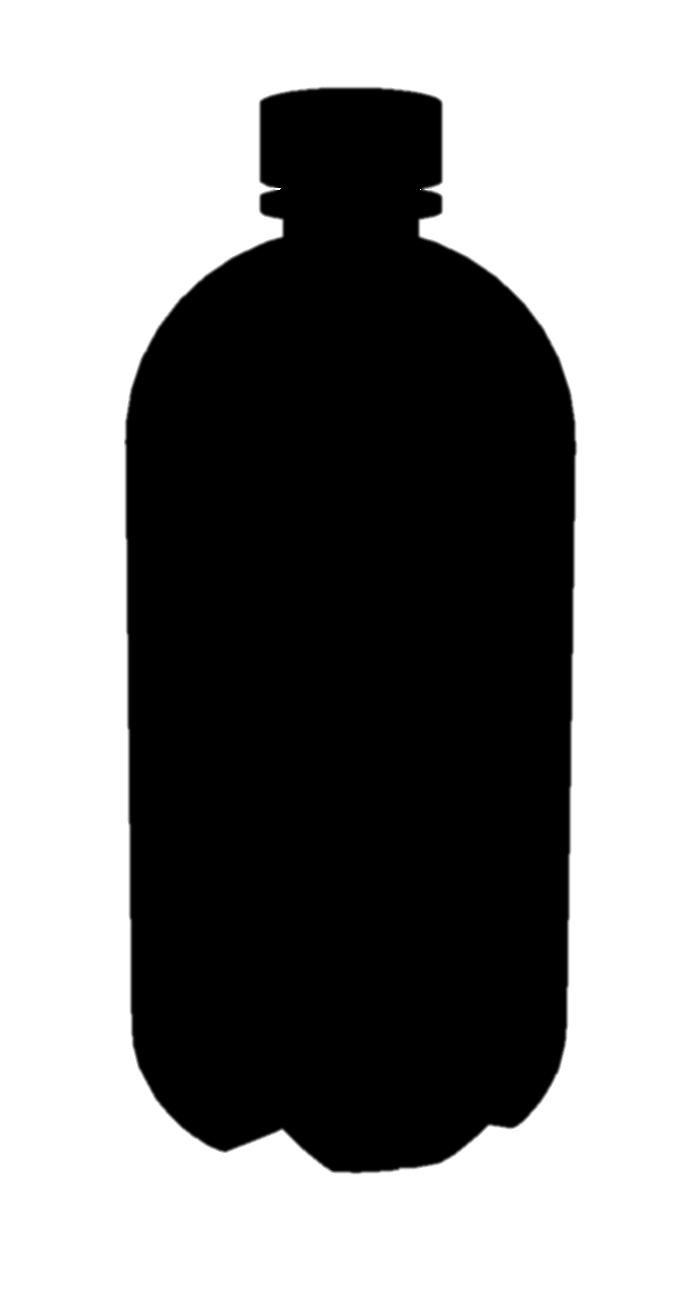}
        };

        \node[below of=gt, node distance=2cm, text width=1cm,
            text centered] {
            \sffamily{Ground truth}
        };

        \draw[-latex, line width=1.25pt, draw=teal] (gt) ++(-0.75, .25) --
            node[midway, above]{$\partial\mathcal{L}$} ++(-1.5, 0)
            (rendering);

        \draw[-latex] (rendering) ++(0.75, -.20) -- node[midway, below]{
            $\mathcal{L}$} ++(1.5, 0) (gt);
    \end{tikzpicture}
    \caption{
        Differentiable renderer preserving the forward rendering link allowing
        backpropagation
    }
    \label{fig:differentiable-rendering}
\end{figure}

Backpropagation is computationally possible if a traceable link of gradients
from the source to the destination and vice versa can be established. With the
conventional graphics rendering pipeline, this link is lost. Differentiable
rendering is a technique that preserves this link
(\cref{fig:differentiable-rendering}). The first known general-purpose
differentiable rendering was proposed in 2014 \cite{loper2014opendr}. Since
then, numerous variants for various use cases have been
developed.\cite{kato2020differentiable}

Differentiable rendering allows a 3D model to be iteratively rendered and
optimised for its parameters. This method has been applied to estimate camera
parameters and volumetric fitting. We apply it to estimate object pose.

  \section{Method}

The proposed method consists of a trainable module appended with a fine tuning
unit. The former is a single-view object localisation and pose prediction unit
followed by differentiable rendering refinement unit (\cref{sec:refinement}).

\subsection{Refinement}
\label{sec:refinement}

We estimate the 3D pose and location of an object in the real world as  a 3D
homography problem with two pieces of information: images from multiple views of
a scene and the camera's (extrinsic) parameters at each viewpoint. The pose and
location estimation is done with differentiable rendering.

\subsubsection{Parameters estimated}

The full pose estimation of an object includes nine degrees of freedom, \viz its
placement in the world, or the \textit{translation vector} $\boldsymbol{t} =
\begin{bmatrix} t_x & t_y & t_z \end{bmatrix}^\top$; its size, or the
\textit{scale vector} $\boldsymbol{s} = \begin{bmatrix} s_x & s_y & s_z
\end{bmatrix}^\top$; and its orientation, or the \textit{rotation matrix}
$\boldsymbol{R}$ composed of the $\mathrm{SO}(3)$ rotation angles.

\subsubsection{Problem formulation and optimisation}

Consider an object captured from two known viewpoints in the world,
$\boldsymbol{E}_1$ and $\boldsymbol{E}_2$. These viewpoints describe the
camera's (extrinsic) parameters. Let the images so captured be denoted
$\boldsymbol{I}_1 \in \mathbb{R}^2$ and $\boldsymbol{I}_2 \in \mathbb{R}^2$,
respectively. As set out in the introduction to this paper in
\cref{sec:introduction}, the reason for using images from two (or more)
viewpoints is to obtain sufficient constraints that are necessary for depth
estimation.

Let $\boldsymbol{v} \in \mathbb{R}^3$ be a set of 3D vertices that form the 3D
CAD model of the object. Now, with a differentiable rendering function $f :
\mathbb{R}^3 \mapsto \mathbb{R}^2$, the CAD model $\boldsymbol{v}$ is rendered
from the aforementioned two viewpoints onto two canvasses, denoted
$\hat{\boldsymbol{I}}_1$ and $\hat{\boldsymbol{I}}_2$, respectively. The 3D mesh
so rendered is initially positioned at some random pose $\hat{\boldsymbol{\Phi}}
= \begin{bmatrix}\hat{\boldsymbol{s}}\hat{\boldsymbol{R}} \vert
    \hat{\boldsymbol{t}}\end{bmatrix}$ in the world. Our goal is to estimate the
    actual pose $\boldsymbol{\Phi} = \begin{bmatrix}\boldsymbol{s}\boldsymbol{R}
    \vert \boldsymbol{t}\end{bmatrix}$, as it is in the world coordinate system.

We could denote the above setup as $ \hat{\boldsymbol{I}}_1 :=
f(\hat{\boldsymbol{\Phi}}; \boldsymbol{v}, \boldsymbol{E}_1)$ and
$\hat{\boldsymbol{I}}_2 := f(\hat{\boldsymbol{\Phi}}; \boldsymbol{v},
\boldsymbol{E}_2)$. Or, more generally, for any viewpoint $n$, \begin{equation}
\label{eqn:rendering-from-viewpoint} \hat{\boldsymbol{I}}_n :=
f(\hat{\boldsymbol{\Phi}}; \boldsymbol{v}, \boldsymbol{E}_n) \end{equation}

\begin{figure}[tb!]
    \centering
    \begin{tikzpicture}
        \node (image) at (0,0) {
            \includegraphics[scale=0.23]{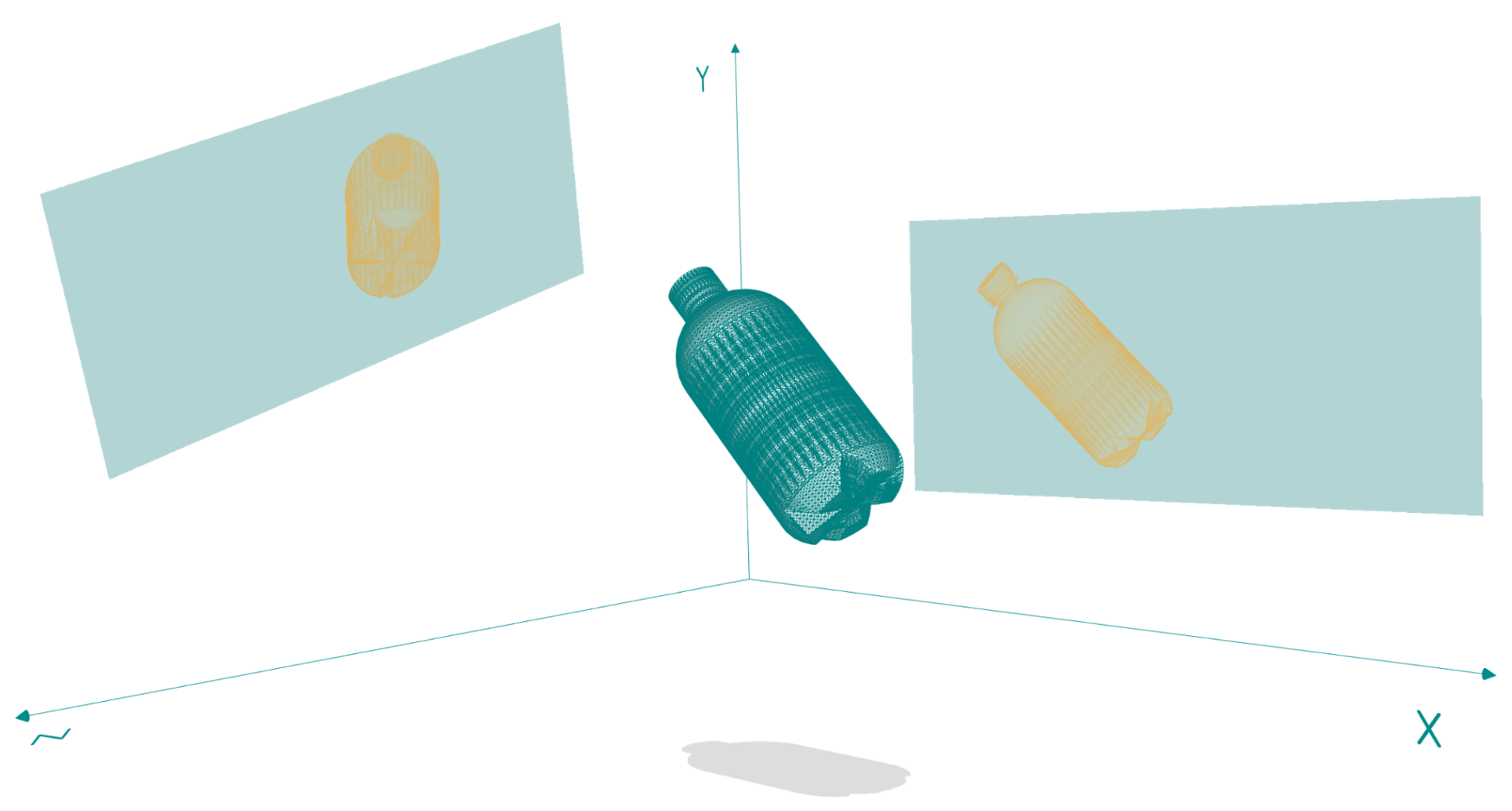}
        };

        \node at (0.25, -0.8) {$\boldsymbol{\Phi}$};
        \node at (2, -0.25) {$\boldsymbol{I}_1$};
        \node at (-1.875, 0.375) {$\boldsymbol{I}_2$};

        \node at (3, 1.25) {$\boldsymbol{E}_1$};
        \node at (-3, 1.3) {$\boldsymbol{E}_2$};
    \end{tikzpicture}
    \caption{
        3D homography: two images $\boldsymbol{I}_1$ and $\boldsymbol{I}_2$ of
        an object are captured from two viewpoints $\boldsymbol{E}_1$ and
        $\boldsymbol{E}_2$, respectively. From this information, the object is
        rendered on both canvasses jointly until the object's renditions
        coincide with its projections in the corresponding original images. This
        yields us the object's pose and location in the world,
        $\boldsymbol{\Phi}$. This illustration shows two viewpoints. The method
        is generalisable to any number of viewpoints.
    }
    \label{fig:method}
\end{figure}
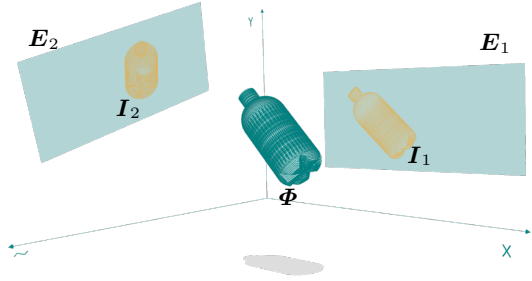

This setup is illustrated in \cref{fig:method}. Note that the estimated object
pose $\hat{\boldsymbol{\Phi}}$ (or the original pose $\boldsymbol{\Phi}$) in the
world coordinate system is, of course, the same for all viewpoints. The many
viewpoints provide good constraints to more accurately deduce the depth of the
object, $t_z$, thus the scale $\boldsymbol{s}$, jointly resolving the
depth-scale ambiguity.

\paragraph{Objective} The rendering $\hat{\boldsymbol{I}}_1$ is compared with
its ground truth $\boldsymbol{I}_1$, and $\hat{\boldsymbol{I}}_2$ is compared
with its ground truth $\boldsymbol{I}_2$. Any discrepancy is backpropagated to
refine the initial estimates $\hat{\boldsymbol{\Phi}}$. This is where
differentiable rendering comes in handy: \begin{equation}
\label{eqn:preliminary-objective}
\frac{\dx}{\dx{\hat{\boldsymbol{\Phi}}}}\left(\hat{\boldsymbol{I}} \circ
\boldsymbol{I}\right) = 0 \end{equation} where ``$\circ$'' is an apt comparison
operator defined by a loss function.

Choosing the right comparison metric that can nudge the CAD model to align with
the ground truth is key to the success of this optimisation step. Hausdorff
distance \cite{aspert2002mesh} $\mathcal{L}_\textrm{H}$ acts as an objective to
minimise the error between the contours of the predicted rendered poses and the
target. However, a rendering beyond the canvas will result in no Hausdorff loss.
To ensure that the estimated poses remain well within the canvas, the
intersection over union (IoU) loss $\mathcal{L}_\textrm{IoU}$ is used.

The combined loss function is given by
\begin{equation}
    \label{eqn:composite-terms}
    \mathcal{L}(\boldsymbol{I}, \boldsymbol{E}; \hat{\boldsymbol{\Phi}},
      \boldsymbol{v}) =
      \lambda_1\mathcal{L}_\textrm{IoU}(\hat{\boldsymbol{I}}, \boldsymbol{I})
      + \lambda_2\mathcal{L}_\textrm{H}(\hat{\boldsymbol{I}}, \boldsymbol{I})
\end{equation}
The coefficients $\lambda_i$ control the importance of each loss term.

Plugging \cref{eqn:composite-terms} in \cref{eqn:preliminary-objective}, we have
$\frac{\dx}{\dx{\hat{\boldsymbol{\Phi}}}}\mathcal{L}(\boldsymbol{I},
\boldsymbol{E}; \hat{\boldsymbol{\Phi}}) = 0$, or the computationally achievable
objective $\boldsymbol{\Phi}^\ast = \argmin_{\hat{\boldsymbol{\Phi}}}
\mathcal{L}(\boldsymbol{I}, \boldsymbol{E}; \hat{\boldsymbol{\Phi}}) $. For
multiview optimisation, this expression is rewritten as
\begin{equation}
    \label{eqn:multiview-objective}
    \boldsymbol{\Phi}^\ast =
    \argmin_{\hat{\boldsymbol{\Phi}}} \frac{1}{N}\sum_{n=1}^{N}\mathcal{L}
    (\boldsymbol{I}_n, \boldsymbol{E}_n; \hat{\boldsymbol{\Phi}})
\end{equation} where $N \ge 2$ is the number of viewpoints from which the
scene has been captured. In classical techniques, a single-step optimisation
strategy such as Newton's method were used.

\subsubsection{Coarse-to-fine orientation optimisation}
\label{sec:search-space-reduction}

Optimisation for all possible orientation Euler angles is a vast space of
$360^\circ \times 360^\circ \times 360^\circ$. To reduce this search space, it
is divided into $k$ bins along each axis, resulting in a reduced $k^3$ search
space. This coarse level yields a few best poses, or a few best bins, $n$ of
which are further pursued to a finer detail \idest an initial $k^3$ searches and
then $\frac{360^3}{k^3}$ for the $n$ best bins, totalling up to $k^3 + n
\left(\frac{360}{k}\right)^3$.

  \section{Experiments}

\subsection{Evaluation metrics}

\subsubsection{Pose estimation using ADD}

The average distortion distance (ADD) between the estimated pose and the ground
truth on an object $\mathcal{M}$ is a frequently used metric to report the
accuracy of pose estimation in 3D space, given by
$\frac{1}{m}\sum_{\boldsymbol{x} \in \mathcal{M}}\Vert
(\boldsymbol{R}\boldsymbol{x} + \boldsymbol{t}) -
(\tilde{\boldsymbol{R}}\boldsymbol{x} + \tilde{\boldsymbol{t}}) \Vert_2$, where
$\boldsymbol{R}$ and $\boldsymbol{t}$ are the ground truth pose, and
$\tilde{\boldsymbol{R}}$ and $\tilde{\boldsymbol{t}}$ are the predicted pose. It
computes the average Euclidian distance between the estimation and the ground
truth. For discrete objects, this metric is the object's centroid. For symmetric
objects, due to the ambiguity arising between points for certain views, the
average closest point distance (ADD-S) \cite{hinterstoisser2012model}, given by
$\frac{1}{m}\sum_{\boldsymbol{x}_1 \in \mathcal{M}}\argmin_{\boldsymbol{x}_2 \in
\mathcal{M}} \Vert (\boldsymbol{R}\boldsymbol{x}_1 + \boldsymbol{t}) -
(\tilde{\boldsymbol{R}}\boldsymbol{x}_2 + \tilde{\boldsymbol{t}}) \Vert_2$, is
adopted. Several seminal and recent works \cite{tremblay2018deep,wu2020deep}
evaluate their methods using ADD and ADD-S. Any grasp pose within a threshold of
ADD or ADD-S is deemed correct.

\subsubsection{Grasp pose rectangle}

We evaluate our method using the \textit{grasping rectangle} metric. It was
designed and published in 2011 \cite{jiang2011efficient} to formalise and
evaluate grasp pose estimations and is used as a standard metric for evaluating
grasp pose. Several state-of-the-art works
\cite{ainetter2021end,luo2020grasp,chu2018deep,asif2018graspnet} evaluate their
methods using grasping rectangle. More advanced and effective metrics have been
developed, a recent one of which \cite{tan2021formulation} accounts for varying
units and scales of features describing the object of interest.

The grasping rectangle metric is a binary decision function which, given a grasp
pose, evaluates it as correct or incorrect. A proposed grasp pose is deemed
correct if the proposed angle falls within a certain threshold angle ($<
30^\circ$) from the ground truth and the proposed grasp meets a certain
threshold intersection-over-union ($> 25\%$) from the grouth truth.

\subsection{Dataset}

\subsubsection{OCID Grasp}

Object Clutter Indoor Dataset (OCID)\cite{suchi2019easylabel} is a dataset
containg 96 cluttered scenes, 89 different objects, and and over 2k point cloud
annotations.

The OCID Grasp dataset\cite{ainetter2021end} is created from OCID by manually
annotating subsets of the latter. These annotations include an object class
label and a corresponding grasp candidate. OCID Grasp consists of 1763 RGB-D
images filtered from OCID, and contains over 11.4k segmented masks and over 75k
grasp candidates.

\subsection{Setup}

We train and experiment our method in the PyTorch environment. 3D rendering and
differentiable rendering are achieved with PyTorch 3D.\cite{ravi2020pytorch3d}

\subsection{Results}

We evaluate object pose estimation by computing the amount of overlap between
the object projected with the estimated pose against the ground truth.
\cref{table:scannet} shows the results obtained by computing the IoU under three
conditions, \viz at least 25\%, 50\%, and 75\% overlaps.

\begin{table}[th!]
    \centering
    \caption{Precision/recall/F1 measures (tested on the ScanNet dataset)}
    \label{table:scannet}
    \begin{tabular}{cccc}
        \toprule
        \textbf{Method}                     & \textbf{IoU $> 0.25$}                     & \textbf{IoU $> 0.5$}              & \textbf{IoU $> 0.75$}     \\
        \midrule
        ODAM \cite{li2021odam}              & 64.7/58.6/61.5                            & 31.2/28.3/29.7                    & 3.8/3.5/3.6               \\
        \midrule
        Vid2CAD \cite{maninis2022vid2cad}   & 56.9/55.7/56.3                            & 34.2/\textbf{33.5}/\textbf{33.9}  & \textbf{10.7/10.4/10.5}   \\
        \midrule
        Ours                                & \textbf{64.9}/\textbf{59.6}/\textbf{62.0} & \textbf{34.7}/31.6/29.2           & 9.6/9.3/8.4               \\
        \bottomrule
    \end{tabular}
\end{table}

\begin{table}[ht!]
    \centering
    \caption{
        Quantitative evaluation on the ScanNet dataset. F1 measures by category
    }
    \label{tab:f1-scannet}
    \begin{tabular}{ccc}
        \toprule
            \textbf{Category}   & \textbf{Single frame} & \textbf{Multiple (3) frames}  \\
            \midrule
            bathtub             & 22.4                  & 62.3                          \\
            bookshelf           & 13.5                  & 51.8                          \\
            cabinet             & 26.3                  & 48.6                          \\
            chair               & 27.4                  & 51.4                          \\
            sofa                & 24.3                  & 48.7                          \\
            table               & 15.4                  & 51.4                          \\
            dustbin             & 27.9                  & 58.7                          \\
            others              & 23.4                  & 47.4                          \\
            \midrule
            global avg.         & 23.3                  & 56.7                          \\
        \bottomrule
    \end{tabular}
\end{table}

\subsubsection{Quantitative evaluation}
\label{sec:quantitative}

We apply the method on the Kinova Jaco arm to examine real-time application.
Since multiple views are not readily available in the real-time setting, the
robot arm makes discrete stops around an object of interest at those viewpoints.

\begin{figure}[ht!]
    \centering
    \begin{tikzpicture}[scale=0.7]
        \begin{axis}[
            ybar,
            bar width=15pt,
            xtick distance=0.5,
            xlabel={Number of frames},
            ylabel={
                $\mathcal{L}(\boldsymbol{I}, \boldsymbol{E};
                \hat{\boldsymbol{\Phi}}, \boldsymbol{v})$
            },
            enlarge x limits={abs=0.5},
            ymin=0,
            scaled ticks=false,
            xtick style={/pgfplots/major tick length=0pt,},
            xtick={1,2,3,4,5}
        ]

        \addplot [error bars/.cd, y dir=both, y explicit relative] coordinates {
            (1,0.9) +- (0,0.082)
            (2,0.4) +- (0,0.074)
            (3,0.12)  +- (0,0.05)
            (4,0.08)  +- (0,0.04)
            (5,0.065) +- (0,0)
        };
        \end{axis}
    \end{tikzpicture}
    \caption{
        Loss convergence for different number of frames. It also shows that the
        error range sharpens as the number of frames increases
    }
    \label{fig:optimisation}
\end{figure}
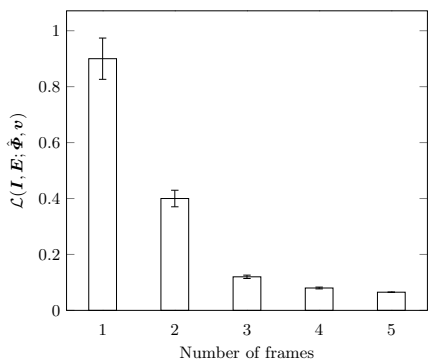

\cref{tab:f1-scannet} shows the improvement of accuracy (by F1) in segmentation
by comparing against the number of frames (one versus three). Pose refinement
through differentiable rendering increases with more time frames
(\cref{fig:optimisation}) and accepts a variable number of frames. We observe
that after a certain threshold on the number of frames, there is very little
information to be gained from other viewpoints, and the convergence rate
stabilised after a few distinct frames.

\cref{table:eucdist} shows a simpler Euclidean distance metric. On a set of five
distinct objects, several sets of trials were conducted with the Kinova Jaco
arm. The distance is scaled to be between 0 to 1, 0 implying pinpoint accuracy
and 1 being no movement of the robot arm.

\begin{table}[hbtp!]
    \centering
    \caption{
        Euclidian distance of the predicted and reached location in world space
        with respect to the base of the robot. These figures are computed by
        normalising the distance between the robot hand and the object over the
        distance between the object and the robot base; 0 signifies no gap
        between the prediction and the ground truth, hence the best estimation;
        1 means the arm has effectively not moved
    }
    \label{table:eucdist}
    \begin{tabular}{cccccc}
        \toprule
        \textbf{Trials} & \textbf{Sphere}   & \textbf{Cube} & \textbf{Banana}   & \textbf{Coffeemug}    & \textbf{bowl} \\
        \midrule
        50              & 0.12              & 0.12          & 0.17              & 0.14                  & 0.16          \\
        75              & 0.11              & 0.07          & 0.14              & 0.12                  & 0.16          \\
        100             & 0.09              & 0.03          & 0.12              & 0.09                  & 0.18          \\
        \bottomrule
    \end{tabular}
\end{table}

\subsection{Object pose estimation for grasping}

To determine the cooperation of this refinement unit with grasping, we predict
object pose on the OCID Grasp dataset and the grap candidates are directly
selected from the annotations. \cref{table:ga} is a sanity check ensuring that
with an ideal candidate proposal and grasp algorithm, the object pose estimation
produces desriable results.

\begin{table}[ht!]
    \centering
    \caption{
        To verify the realism of object pose estimation, we apply the grasp pose
        angles directly obtained from the annotations in the OCID Grasp dataset
        after transforming to the camera coordinates. Accurate object pose
        estimation results in a successful grasp. Conversely, a successful grasp
        implies accurate object pose estimation.
    }
    \label{table:ga}
    \begin{tabular}{ccc}
        \toprule
        \textbf{Dataset}    & \textbf{Grasp accuracy}(\%)   & \textbf{IoU}  \\
        \midrule
        OCID Grasp          & 99.65\%                       & 99.32\%       \\
        \bottomrule
    \end{tabular}
\end{table}

  \section{Conclusion}

In this work, we present an optimisation strategy for object pose estimation by
refining coarse estimations through multi-view differentiable rendering. This
approach avoids rich sensor data such as point clouds or other depth data and
relies on RGB images and camera parameters at different viewpoints. This
approach is comparable to or outperforms the state of the art under different
conditions. We experiment this method to evaluate pose estimation depending on
the grasp success rate by applying annotated grasp candidates. This method can
be augmented to any well-performing segmentation model and prepended to any
grasp candidate estimation algorithm.

\textbf{Limitation.} Since the refinement module is an online optimisation unit,
pose estimation optimisation happens right at inference time. This is a
limitation on time-sensitive application.

\textbf{Acknowledgements.} This material is based upon work supported by the
National Science Foundation under Grant No. CMMI 1826258.

  \clearpage
  \bibliographystyle{splncs04}
  \bibliography{references}
\end{document}